%% file: main.tex
\documentclass[preprint,journal]{vgtc}       

\usepackage{mathptmx}
\usepackage{graphicx}
\usepackage{subfigure} 
\usepackage{times}
\usepackage{amsmath}
\usepackage{enumitem}

\usepackage[bookmarks,backref=true,linkcolor=black]{hyperref} 
\hypersetup{
  pdfauthor = {},
  pdftitle = {},
  pdfsubject = {},
  pdfkeywords = {},
  colorlinks=true,
  linkcolor= black,
  citecolor= black,
  pageanchor=true,
  urlcolor = black,
  plainpages = false,
  linktocpage
}

\onlineid{1290}

\vgtccategory{Algorithm/Technique}

\vgtcinsertpkg

\title{DECE: Decision Explorer with Counterfactual Explanations for Machine Learning Models}

\author{Furui Cheng, Yao Ming, Huamin Qu}
\authorfooter{
\item
 Furui Cheng, and Huamin Qu are with Hong Kong University of Science and Technology. E-mail: \{fchengaa, huamin\}@ust.hk.
\item
 Yao Ming is with Bloomberg L.P. This work was done when he was at Hong Kong University of Science and Technology. E-mail: yming7@bloomberg.net
}

\usepackage{xcolor}
\usepackage{tikz}
\usepackage{dsfont}

\newcommand{\tikzcircle}[2][red,fill=red]{\tikz[baseline=-0.5ex]\draw[#1,radius=#2] (0,0) circle ;}%

\definecolor{lightOrange}{RGB}{252,196,166}
\definecolor{darkOrange}{RGB}{244,114,54}

\newcommand{\inpoint}{\tikzcircle[lightOrange,fill=lightOrange]{2pt}}
\newcommand{\outpoint}{\tikzcircle[darkOrange,fill=darkOrange]{2pt}}

\newcommand{\system}{{DECE}}
\newcommand{\subsetCF}{{r-counterfactuals}}

\DeclareMathOperator*{\argmin}{argmin}
\DeclareMathOperator*{\argmax}{argmax}

\usepackage[normalem]{ulem} 

\newcommand{\revise}[2]{{#2}}

\newcommand{\del}[1]{}

\newcommand{\etal}{\textit{et al.}}
\newcommand{\eg}{\textit{e.g.}}
\newcommand{\ie}{\textit{i.e.}}

\abstract{
\input{content/abstract.tex}
} 


\teaser{
 \centering
 \includegraphics[width=16cm]{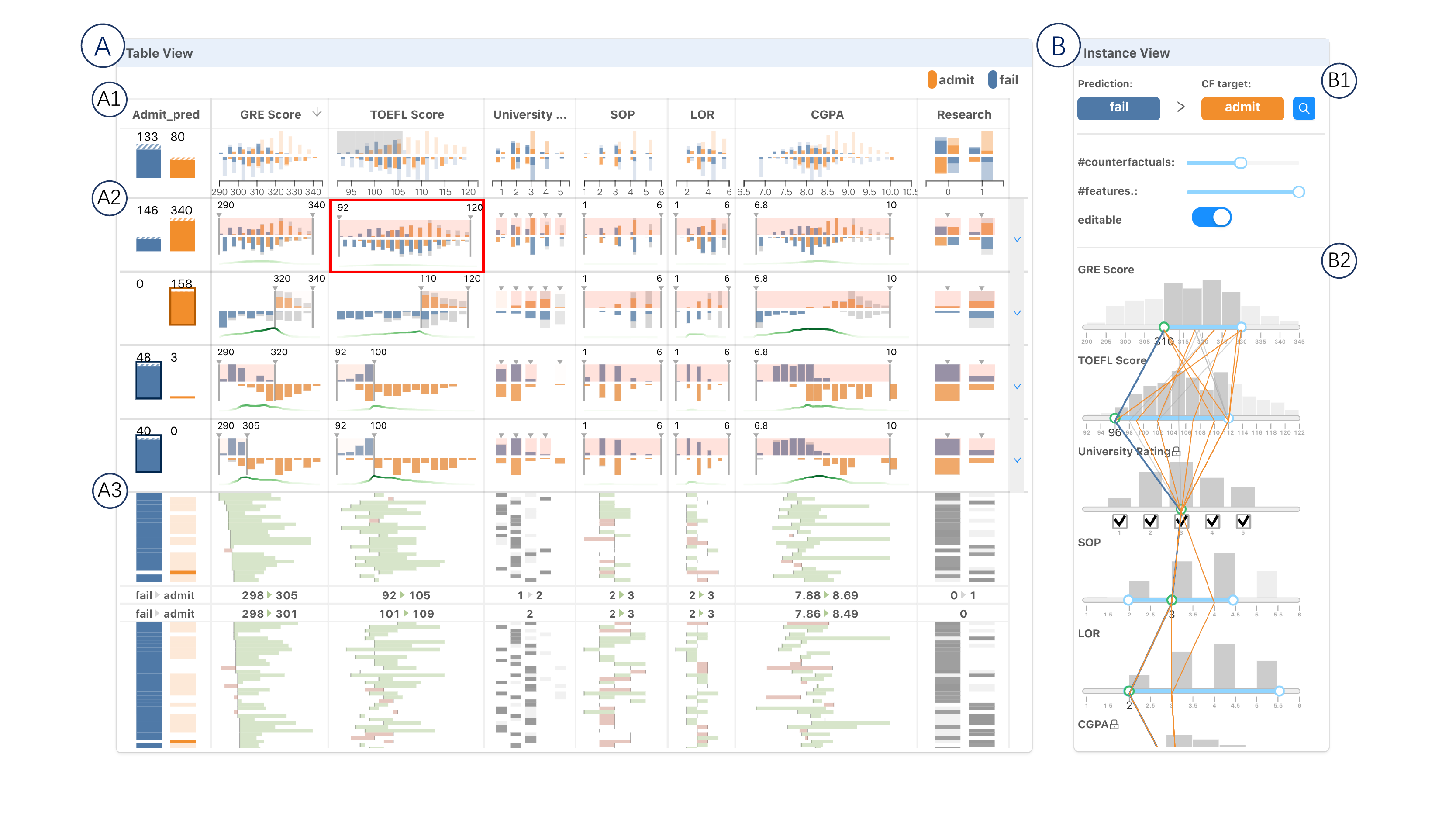}
      \vspace{-0.35in}

  \caption{The \system{} interface for exploring a machine learning model's decisions with counterfactual explanations. 
  The user uses the \textit{table view} (A) for subgroup level analysis. 
  The table header (A1) supports the exploration of the table with sorting and filtering operations.
  The subgroup list (A2) presents the subgroups in rows and summarizes their counterfactual examples.
  The user can interactively create, update, and delete a list of subgroups.
  The instance lens (A3) visualizes each instance in the focused subgroup as a single thin horizontal line.
  In the \textit{instance view} (B), the user can customize (B1) and inspect the diverse counterfactual examples of a single instance in an enhanced parallel coordinate view (B2). 
  }
  \label{fig:teaser}
}

\begin{document}



\maketitle


\input{content/introduction.tex}

\input{content/related-work.tex}

\input{content/design-requirement.tex}

\input{content/methodology.tex}

\input{content/system.tex}

\input{content/evaluation.tex}

\input{content/discussion.tex}

\bibliographystyle{abbrv}
\bibliography{bibliography}
\end{document}

%% file: content/abstract.tex
%
%
%
With machine learning models being increasingly applied to various decision-making scenarios, people have spent growing efforts to make machine learning models more transparent and explainable. Among various explanation techniques, \textit{counterfactual explanations} have the advantages of being human-friendly and actionable --- a counterfactual explanation tells the user how to gain the desired prediction with minimal changes to the input. Besides, counterfactual explanations can also serve as efficient probes to the models' decisions.
In this work, we exploit the potential of counterfactual explanations to understand and explore the behavior of  machine learning models. We design \system, an interactive visualization system that helps understand and explore a model's decisions on individual instances and data subsets, supporting users ranging from decision-subjects to model developers. 
{\system} supports exploratory analysis of model decisions by combining the strengths of counterfactual explanations at instance- and subgroup-levels.
We also introduce a set of interactions that enable users to customize the generation of counterfactual explanations to find more actionable ones that can suit their needs.
Through three use cases and an expert interview, we demonstrate the effectiveness of {\system} in supporting decision exploration tasks and instance explanations.

%% file: content/introduction.tex
\section{Introduction}


In recent years, we have witnessed an increasing adoption of machine learning (ML) models to support data-driven decision-making in various application domains, which include decisions on loan approvals, risk assessment for certain diseases, and admissions to various universities. Due to the complexity of these real-world problems, well-fitted ML models with good predictive performance often make decisions via complex pathways, and it is difficult to obtain human-comprehensible explanations directly from the models. The lack of interpretability and transparency could result in hidden biases and potentially harmful actions, which may hinder the real-world deployment of ML models. \looseness=-1

To address this challenge, a variety of post-hoc model explanation techniques have been proposed \cite{iml_cacm_review}. Most of the techniques explain the model's decisions by calculating feature attributions or through case-based reasoning. An alternative approach for providing human-friendly and actionable explanations is to present users with \textit{counterfactuals}, or \textit{counterfactual explanations} \cite{lewis2013counterfactuals, wachter2017counterfactual}. 
The method answers this question: How does one obtain an alternative or desirable prediction by altering the data just a little bit? For instance, a person submitted a loan request but got rejected by the bank based on the recommendations made by an ML model. Counterfactuals provide explanations like ``if you had an income of \$40 000 rather than \$30 000, or a credit score of $700$ rather than $600$, your loan request would have been approved.'' Counterfactual explanations are user-friendly to the general public as they do not require prior-knowledge on machine learning \cite{binns2018s}. 
\revise{Counterfactual-based explanations also have real-world implications since they provide actionable recommendations to change the outcome of a model. }{
Another advantage is that counterfactual explanations are not based on approximation but always give exact predictions by the model \cite{mothilal2020explaining}. Watcher \etal\ summarize the three important scenarios for decision subjects as understanding the decision, contesting the (undesired) decision, and providing actionable recommendations to alter the decision in the future \cite{wachter2017counterfactual}.
}
For model developers, counterfactual explanations can be used to analyze the decision boundaries of a model, which can help detect the model's possible flaws and biases \cite{wexler2019if}. For example, if the counterfactual explanations for loan rejections all require changing, \eg, the gender or race of an applicant, then the model is potentially biased.  \looseness=-1

Recently, a variety of techniques have been developed to generate counterfactual explanations \cite{wachter2017counterfactual}. However, most of the techniques focus on providing explanations for the prediction of individual instances \cite{wexler2019if}. To examine the decision boundaries and analyze model biases, the corresponding technique should be able to provide an overview of the counterfactual examples generated for a population or a selected subgroup of the population. Furthermore, in real-world applications, certain constraints are needed such that the counterfactual examples generated are feasible solutions in reality. For example, one may want to limit the range of credit score changes when generating counterfactual explanations for a loan application approval model. \looseness=-1

An interactive visual interface that can support the exploration of the counterfactual examples to analyze a model's decision boundaries, as well as edit the constraints for counterfactual generation, can be extremely helpful for ML practitioners to probe the model's behavior and also for everyday users to obtain more actionable explanations. Our goal is to develop a visual interface that can help model developers and model users understand the model's predictions, diagnose possible flaws and biases, and gain supporting evidence for decision making. We focus on ML models for classification tasks on tabular data, which is one of the most common real-world applications. The proposed system, \emph{Decision Explorer with Counterfactual Explanations} (\system),  supports interactive subgroup creation from the original data-set and cross-comparison of their counterfactual examples by extending the familiar tabular data display. This greatly eases the learning curve for users\revise{}{ with basic data analysis skills}. More importantly, since analyzing decision boundaries for models with complex prediction pathways is a challenging task, we propose a novel approach to help users interactively discover simple yet effective decision rules by analyzing counterfactual examples. An example of such a rule is `` Body Mass Index (BMI) below 30 (almost) ensures that the patient does not have diabetes, no matter how the other attributes of the patient change.'' By searching for the corresponding counterfactual examples, we can verify the robustness of such rules. To ``flip'' the prediction given in this example, the BMI of a diabetic patient must be above 30. Such rules can be presented to the domain experts to help validate a model by checking if they align with domain knowledge. Sometimes new insights are gained from the identified rules.  \looseness=-1

To summarize, our contributions include:
\begin{itemize}
    \vspace{-0.1in}
    \setlength\itemsep{-0.4em}
    \item \system{}, a visualization system that helps model developers and model users explore and understand the decisions of ML models through counterfactual explanations.  \looseness=-1
    \item \revise{An interactive visual interface}{A subgroup counterfactual explanation method} that supports exploratory analysis and hypothesis refinement using subgroup counterfactual explanations.
    \looseness=-1
    \item Three use cases and an expert interview that demonstrate the effectiveness of our system.
\end{itemize}

%% file: content/related-work.tex
\section{Related Work}



\subsection{Counterfactual Explanation}\label{sec:related-cf}

Counterfactual explanations aim to find a minimal change in data that ``flips'' the model's prediction. They provide actionable guidance to end-users in a user-friendly way. The use of counterfactual explanations is supported by the study of social science \cite{miller2019explanation} and philosophy literature \cite{lewis2013counterfactuals, ruben2015explaining}.  \looseness=-1

Wachter \etal\ \cite{wachter2017counterfactual} first proposed the concept of unconditional counterfactual explanations and a framework to generate counterfactual explanations by solving an optimization problem. With a user-desired prediction $y'$ that is different from the predicted label $y$, a counterfactual example $c$ against the original instance $x$ can be found by solving
\begin{align}
    \label{eq:wachter}
    \argmin_{c} \max_\lambda \lambda(f_w (c)-y')^2 + d(x, c),
\end{align}
where $f_w$ is the model and $d(\cdot,\cdot)$ is a function that measures the distance between the counterfactual example $x'$ to the original instance $x$.
Ustun \etal\ \cite{ustun2019actionable} further discussed factors that affected the feasibility of counterfactual examples and designed an integer programming tool to generate diverse counterfactuals to linear models. Russell \cite{russell2019efficient} designed a similar method to support complex data with mixed value (a contiguous range or a set of discrete special value). Lucic \etal designed DATE \cite{lucic2019actionable} to generate counterfactual examples to non-differentiable models with a focus on tree ensembles. Mothilal \etal\ proposed a quantitative evaluation framework and designed DiCE \cite{mothilal2020explaining}, a model-agnostic tool to generate diverse counterfactuals. Karimi \etal\ \cite{karimi2019model} proposed a general pipeline by solving a series of satisfiability problems. Most existing work focuses on the generation and evaluation of the counterfactual explanations \cite{Ghazimatin2020PRINCE, madumal2019explainable, singla2019explanation}.

Instead of generating counterfactual explanations, our work attempts to solve the question of \textit{how to convey counterfactual explanation information to a subgroup using visualization}. 
Another focus of our work is to design interactions to help users find more feasible and actionable counterfactual explanations, \eg, with a more proper distance measurement suggested by Rudin \cite{rudin2019stop}.  \looseness=-1


\subsection{Visual Analytics for Explainable Machine Learning}

A variety of visual analytics techniques have been developed to make machine learning models more explainable. Common use cases for the explainable techniques include understanding, diagnosing, and evaluating machine learning models. Recent advances have been summarized in a few surveys and framework papers \cite{Liu:2017:survey,hohman2018visual,spinner:2020:explainer}.  \looseness=-1

Most existing techniques target at providing explainability for deep neural networks. Liu \etal\ \cite{Liu:2017:cnnvis} combined matrix visualization, clustering, and edge bundling to visualize the neurons of a CNN image classifier, which helps developers understand and diagnose CNNs. Alsallakh \etal\ \cite{Alsallakh:2018:cnnHierarchy} developed Blocks to identify the sources of errors of CNN classifiers, which inspired their improvements on CNN architecture. Strobelt \etal\ \cite{Strobelt:2018:lstmvis} studied the dynamics of the hidden states of recurrent neural networks (RNN) as applied to text data using parallel coordinates and heatmaps. Various other work followed this line of research to explain deep neural networks by examining and analyzing their internal representations \cite{Kahng:2018:ActiVis,hohman2020summit,liu:2018:DGM,ming:2017:rnnvis,wang:2018:ganviz,wang:2019:dqnviz,strobelt:2019:seq2seqvis}.  \looseness=-1

The common limitation of these techniques is that they are often model-specific. It is challenging to generalize them to other types of models that emerge as machine learning research advances. In our work, we study counterfactual explanations from a visualization perspective and develop a model-agnostic solution that applies to both instance- and subset-levels.  \looseness=-1

The idea of model-agnostic explanation was popularized in LIME \cite{Ribeiro:2016:WIT}. Visualization researchers have also studied this idea in Prospector \cite{krause2016interacting}, RuleMatrix \cite{ming:2018:rulematrix}, and the What-If Tool \cite{wexler2019if}. 
Closely related to our work, the What-If Tool adopts a perturbation-based method to help users interactively probe machine learning models. It also offers the functionality of finding the nearest ``counterfactual'' data points with different labels. Our work investigates the general concept of counterfactuals that are independent of the available dataset. Besides, we utilize subset counterfactuals to study and analyze the decision boundaries of machine learning models.  \looseness=-1





%% file: content/design-requirement.tex
\section{Design Requirements}
\label{sec:design-requirements}
Our goal is to develop a generic counterfactual-based model explanation tool that helps users get actionable explanations and understand model behavior. For general users like decision subjects, counterfactual examples can help them understand how to improve their profile to get the desired outcome. For users like model developers or decision-makers, we aim to provide counterfactual explanations that can be generalized for a certain group of instances. To reach our goal, we first survey design studies for explainable machine learning to understand general user needs \cite{Ahn:2019:fairsight,hohman2019gamut,Kahng:2018:ActiVis,krause2017workflow,krause2016interacting,Liu:2017:cnnvis,ming:2018:rulematrix, spinner:2020:explainer,Strobelt:2018:lstmvis,wexler2019if}. Then we analyze these user needs considering the characteristics of counterfactual explanations and identify two key levels of user interests that relate to counterfactuals: \emph{instance-level} and \emph{subgroup-level} explanations. 

\revise{}{
Instead of understanding how the model works globally, decision subjects are more interested in knowing how a prediction is made based on an individual instance, like their profile. This makes instance-level explanations more essential for decision subjects.}
At the \textbf{instance-level}, we aim to empower users with the ability to:

\vspace{-0.05in}
\begin{enumerate}[label=\textbf{R\arabic*}]
    \setlength\itemsep{-0.1em}
    
    
    \item\label{r:instance}
    \textbf{Examine the \emph{diverse} counterfactuals to an instance.}
    Accessing an explanation of the model's prediction on a specific instance is a fundamental need. To be more actionable, it is often helpful to provide several counterfactuals that cover a diverse range of conditions than a single closest one \cite{wachter2017counterfactual}.
    The user should also be able to examine and compare them in an efficient manner. The user can examine the different options and choose the best one based on individual needs.
    
    \item\label{r:customize}
    \textbf{Customize the counterfactuals with user-preferences.} 
    Providing multiple counterfactuals and hoping that one of them matches user needs may not always work. In some situations, it is better to allow users to directly specify preferences or constraints on the type of counterfactuals they need. For example, 
    one home buyer may prefer a larger house, while another buyer only cares about the location and neighborhood.
    

\end{enumerate}
\vspace{-0.05in}

Similar to the ``eyes beat memory'' principle, it is hard to view and memorize multiple instance-level explanations and derive an overall understanding of the model. Explaining machine learning models at a higher and more abstract level than an instance can help users understand the general behavior of the model \cite{Kahng:2018:ActiVis, ming:2018:rulematrix}. One of our major goals is to enable subgroup-level analysis based on counterfactuals. Subgroup analysis of the counterfactuals is crucial for users like model developers and policy-makers, who need an overall comprehension of the model and the underlying dataset. A subgroup also provides a flexible scope that allows iterative and comparative analysis of model behavior.  At the \textbf{subgroup-level}, we aim to provide users the ability to: 

\vspace{-0.05in}
\begin{enumerate}[resume*]
    \setlength\itemsep{-0.1em}
    
    \item\label{r:selection}
    \revise{}{\textbf{Select and refine a data subgroup of interest.}
    To conduct subgroup analysis using counterfactual explanations, the users should first be equipped with tools to select and refine subgroups. Interesting subgroups could be those formed from users' prior knowledge or those that could suggest hypotheses for describing the model.
    For instance, a high glucose level is often considered a strong sign of diabetes. The user (patient or doctor) may be interested in a subgroup consisting of low glucose-level patients labeled as healthy, and see if most of their counterfactual examples (patients with diabetes) have high glucose levels. However, drilling down to a proper subgroup (\ie, an appropriate glucose-level range) is not easy. Providing essential tools to create and iteratively refine subgroups could largely benefit users' exploration processes.}
    
    \item\label{r:subgroup}
    \revise{}{\textbf{Summarize the counterfactual examples of a subgroup of instances.} With a subgroup of instances, we are interested in the distribution of their counterfactual examples. Do they share similar counterfactual examples? Are there any counterfactual examples that lie inside the subgroup? An educator would be interested in knowing if the performance of a certain group of students can be improved with a single action. It is also useful for model developers to form and verify their hypothesis by investigating a general prediction pattern over a subgroup.}

    \item\label{r:comparison}
    \textbf{Compare the counterfactual examples of different subgroups.} 
    Comparative analysis across different groups could lead to deeper understanding. It is also an intuitive way to reveal potential biases in the model\del{ or the dataset}. For instance, to achieve the same desired annual income, do different genders or ethnic groups need to take different actions? 
    Comparison can provide evidence for progressive refinement of subgroups, helping users to identify a salient subgroup that has the same predicted outcome.

\end{enumerate}

%% file: content/methodology.tex
\section{Counterfactual Explanation}\label{sec:methodology}




In this section, we first introduce the techniques and algorithms that we use to generate diverse actionable explanations with customized constraints (\ref{r:instance}, \ref{r:customize}). Subsequently, we propose the definition of rule support counterfactual examples, which is designed to support exploring a model's local behavior in the subgroup-level analysis (\ref{r:selection}, \ref{r:subgroup}, \ref{r:comparison}). 

\subsection{Generating Counterfactual Examples}

As introduced in \autoref{sec:related-cf}, given a black box model $f: X \rightarrow Y$, the problem of generating counterfactual explanations to an instance $\mathbf{x}$ is to find a set of examples $\{\mathbf{c}_1, \mathbf{c}_2, ..., \mathbf{c}_k\}$ that lead to a desired prediction $y'$, which are also called counterfactual examples (CF examples). The CF examples can suggest how a decision subject could \emph{act} to achieve the user's targets. The problem we address in this section is how to generate CF examples that are valid and actionable. 

CF examples are actionable when they appropriately consider proximity, diversity, and sparsity.
First, the generated examples should be \emph{proximal} to the original instance, which means only a small change has to be made to the user's current situation. However, one predefined distance metric cannot fit every need because people may have different preferences or constraints \cite{russell2019efficient}. 
Thus, we want to offer \emph{diverse} options (\ref{r:instance}) to choose from and also allow them to add \emph{constraints} (\ref{r:customize}) to reflect their preferences or narrow their searches. Finally, to enhance the interpretability of the examples, we want the examples to be \emph{sparse}, which means that only a few features need to be changed. 

We follow the framework of DiCE \cite{mothilal2020explaining} and design an algorithm to generate both valid and actionable CF examples using three procedures. 
First, we generate raw CF examples by considering their validity, proximity, and diversity. 
To make the trade-off between these three properties, we optimize a three-part loss function as 

\begin{equation}
\label{eq:loss}
L = L_{valid} + \lambda_1 L_{dist} + \lambda_2 L_{div}.
\end{equation}

\textbf{Validity.}
The first term $L_{valid}$ is the validity term, which ensures the generated CF examples reach the desired prediction target. We define it as $$L_{valid} = \sum_{i=1}^{k}\textup{loss}(f(\mathbf{c}_i), y'),$$ in which the $\textup{loss}$ is a metric to measure the distance between the target $y'$ and the prediction of each CF example $f(\mathbf{c}_i)$. 
For classification tasks, we only require that the prediction flips, and high confidence or possibility of the prediction result is not necessary. 
Thus, instead of choosing the commonly used $L_1$ or $L_2$ loss, we let $\textup{loss}$ be the ranking loss with zero margins. In a binary classification task, the loss function is
$
\textup{loss}(y_{pred}, y') = max(0,\, -\,y'*(y_{pred} - 0.5)),
$
in which the target $y' = \pm 1$, and $y_{pred}$ is the prediction of the CF example by the model $f(\mathbf{c})$, which is normalized to $[0, 1]$.

\textbf{Proximity.}
As suggested by the proximity requirement, we want the CF examples to be close to the original instance by minimizing $L_{dist}$ in the loss function. We define the proximity loss as the sum of the distance from the CF examples to the original instance: 
$$L_{dist} = \sum_{i=1}^{k}\textup{dist}(\mathbf{c}_i, \mathbf{x}).$$ 

\revise{}{We choose a weighted Heterogeneous Manhattan-Overlay Metric (HMOM) \cite{wilson1997improved} to calculate the distance as follows:}
\begin{equation}
\label{eq:dist}
\textup{dist}(\mathbf{c}, \mathbf{x}) = \sum_{f\in F}d_{f}(\mathbf{c}^f, \mathbf{x}^f),
\end{equation}
\revise{}{where}
$$
d_{f}(\mathbf{c}^f, \mathbf{x}^f) = \left\{\begin{matrix}
\frac{|\mathbf{c}^f - \mathbf{x}^f|}{(1+MAD_f)\cdot range_f} & \textup{if}\ f\ \textup{indexes a continuous feature}\\ 
\mathds{1}({\mathbf{c}^f \neq  \mathbf{x}^f})  & \textup{if}\ f\ \textup{indexes a categorical feature}
\end{matrix}\right..
$$
\revise{}{For continuous features, we apply a normalized Manhattan distance metric weighted by $1/(1+MAD_f)$ as suggested by Watcher \etal\ \cite{wachter2017counterfactual}, where $MAD_f$ is the median absolute deviation (MAD) value of the feature $f$. By applying this weight, we encourage the feature values with large variation to change while the rest stay close to the original values. 
For categorical features, we apply an overlap metric $\mathds{1}({\mathbf{c}^f \neq  \mathbf{x}^f}$), which is 1 when $\mathbf{c}^f \neq  \mathbf{x}^f$ and 0 when $\mathbf{c}^f = \mathbf{x}^f$.
}

\textbf{Diversity.}
To achieve diversity, we encourage the generated examples to be separated from each other. Specifically, we calculate the pairwise distance of a set of CF examples and minimize:
$$
L_{div} = -\frac{1}{k}\sum_{i=1}^{k}\sum_{j=i}^{k}\textup{dist}(\mathbf{c}_i, \mathbf{c}_j),
$$ where the distance metric is \revise{the same as the one used in the proximity loss term.}{defined in \autoref{eq:dist}.}

To solve the above optimization problem, we could use any gradient-based optimizers. For simplicity, we use the classic stochastic gradient descent (SGD) in this work. 
As discussed in \ref{r:customize}, we want to allow users to specify their preferences by adding constraints in the generation process. The constraints decide if and within what range a feature value should change. To fix the immutable feature values, we update them with a masked gradient, \ie, the gradient to the immutable feature values is set to 0. 
We also run a clip operation every $\textup{K}$ iteration to project the feature values to a feasible value in the range. 

\textbf{Sparsity.}
The sparsity requirement suggests that only a few feature values should change. To enhance the sparsity of the generated CF examples, we apply a \revise{straightforward post-hoc procedure.}{feature selection procedure.} 
We first generate raw CF examples from the previous procedure.
\revise{}{Then we select the top-$k$ features for each CF example separately with the normalized maximum value changes weighted by $1/(1+MAD_f)$.}
At last, we repeat the above optimization procedure with only these $k$ features by masking the gradient of other features.
The generated CF examples are sparse with at most k changed feature values.

\textbf{Post-hoc validity.}
In previous procedures, we treat the value of each continuous feature as a real number. However, in a real-world dataset, features may \revise{}{be integers or} have certain precisions. For example, \revise{a person's age}{a patient's number of pregnancies} should be an integer, and \revise{it does not make sense if we say someone's age is 23.14}{a value with decimals for this feature can bring confusion to users}. Thus, we project each CF example ${\mathbf{c}_i}$ to a meaningful one ${\mathbf{\tilde{c}}_i}$. 
Let the validity of projected CF examples, ${\mathbf{\tilde{c}}_i}$, exist as post-hoc validity. 
We design a post-hoc process as the third procedure to improve the post-hoc validity by refining the projected CF examples. 

In each step of the process, we calculate the gradient of each feature to the loss $L$ (\autoref{eq:loss}),
$grad_i = \nabla {\mathbf{\tilde{c}}_i}\ loss(\mathbf{\tilde{c}}_i, \mathbf{x})$, 
and update the projected CF example by updating the feature value with the largest absolute normalized gradient value $j = \argmax_{f\in F} (| grad_i^{f} | )$:
\label{eq:post-step}
\begin{equation}
    \tilde{c}_{i, t+1}^{j} = \tilde{c}_{i, t}^{j} + \max  (p^{j},\ \epsilon | grad_i^{j}| )\ \textup{sign} ( grad_i^{j}),
\end{equation}
where $p^{j}$ notes the unit of the feature $j$ and $\epsilon$ is a given hyper-parameter, which usually equals the learning rate in the SGD process above. The process ends when the updated CF example is valid, or the number of steps reaches a maximum number, which is often set as the number of features.

\subsection{Rule Support Counterfactual Examples}
\label{sec:subgroup-method}
\del{As discussed in \autoref{sec:design-requirements}, a subgroup-level analysis of model behavior is a major requirement.}
We first propose a subgroup-level exploratory analysis procedure for understanding a model's local behavior. Then we introduce the definition of rule support counterfactual examples (\subsetCF), which is designed to support such an exploratory analysis procedure.

\begin{figure}
    \centering
    \includegraphics[width=0.9\linewidth]{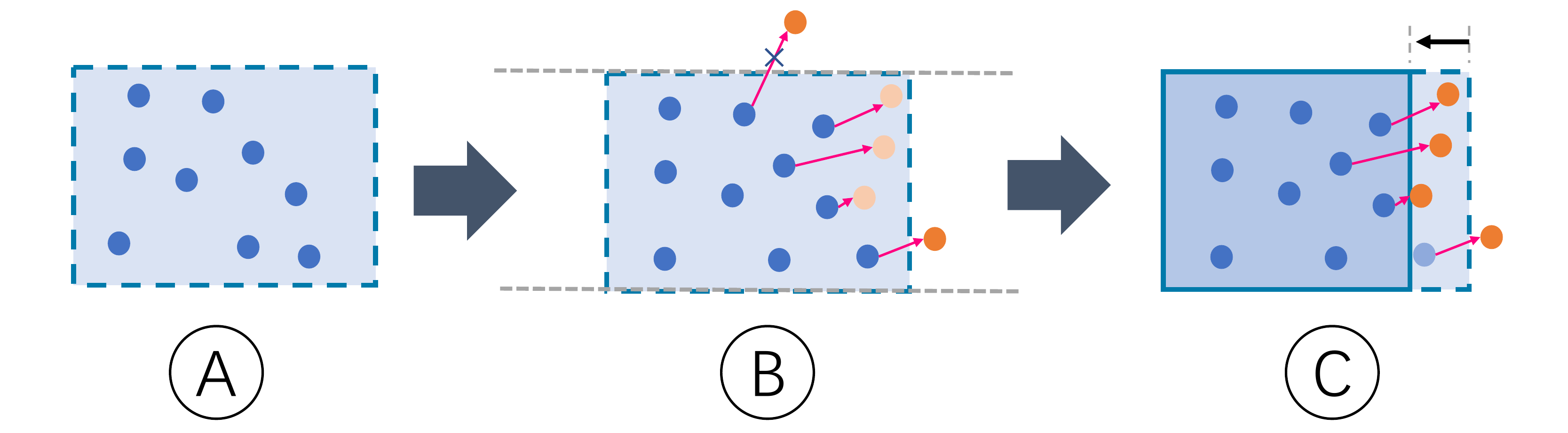}
     \vspace{-0.15in}
    \caption{
    A simple exploratory analysis with \subsetCF{}.
    A. A hypothesis is proposed by selecting a subgroup. 
    B. R-counterfactuals are generated against the subgroup. \inpoint{} are instances within the subgroup, and \outpoint{} are instances outside the subgroup.
    C. The hypothesis is refined to a new subgroup that excludes the previous CF examples. \looseness=-1
    }
    \label{fig:subgroup}
     \vspace{-0.05in}

\end{figure}

One of the major goals of exploratory analysis is to suggest and assess hypotheses \cite{tukey1977eda}. The exploration starts with a hypothesis about the model's prediction on a subgroup proposed by users. 
A hypothesis is an assertion in the form of an if-else rule that describes a model's prediction, \eg, ``\textit{People who are under 30 years old and whose BMI is under 35 will be predicted healthy by the diabetes prediction model.}''
No matter how the other features change\revise{}{ (\eg, smoking or not)}, as long as the two conditions (under 30 years old with BMI under 35) hold, the person is unlikely to have diabetes.
Each hypothesis describes the model's behavior on a subgroup defined by range constraints on a set of features (\autoref{fig:subgroup}A):
 \begin{equation}
     S = D \cap I'_1 \times I'_2 \times ... \times I'_k,
 \end{equation}
where $D$ is the dataset and $I'_j$ defines the value range of feature $j$.  
\revise{}{
The value range is a continuous interval for continuous features, and a set of selectable categories for categorical features.
}

The users expect to find out whether the model's prediction on the collected data conforms to the hypothesis and, more importantly, if the hypothesis generalizes in unseen instances. 
CF examples can be used to answer the two questions. Intuition suggests that if we can find a feasible CF example against one of the instances in the subgroup, the hypothesis might not be  valid. For example, if we can find a person whose prediction for having diabetes can be flipped to positive but age $< 30$ and BMI $ <35$, the hypothesis that ``people under 30 years old with BMI under 35 will be predicted healthy'' does not hold. Otherwise, the hypothesis is supported by the CF examples. 

For an invalid hypothesis, CF examples also suggest how to refine it. For example, if a CF example tells that ``\textit{a 29-year-old \revise{}{smoker} whose BMI is 30 is predicted as diabetic}'', it suggests that the user may narrow the subgroup to $age < 30\ \textup{and}\ BMI < 30$ or refer to other feature values \revise{}{(\eg, $smoking \in \{ no\}$)}. 
With multiple rounds of hypothesis, users can understand the model's prediction on a subgroup of interest. 

In our initial approaches, we find that unconstrained CF examples would overwhelm users due to the complex interplay of multiple features. Thus, we simplify the problem by only focusing on one feature at a time. This is achieved by generating a group of constrained CF examples called rule support counterfactual examples (\subsetCF). These are counterfactuals that support a rule. 
Specifically, with a given subgroup, we generate CF examples by only allowing the value of one feature $j$ to change in the domain $X_j$. In contrast, other feature values can only vary in the limited range, $I'_j$.
For each feature $j$, we generate \subsetCF \ (\autoref{fig:subgroup}B) by solving:
 \begin{equation}
     \texttt{r-counterfactuals}_j : \underset{\{\mathbf{c}_i\}}{\textup{argmin}} \sum_{\mathbf{x}_i\in S} L(\mathbf{x}_i, \mathbf{c}_i),
\end{equation}
\begin{equation}
     \textup{such that:}\ \mathbf{c}_i \in I'_1 \times I'_2 \times ... \times X_j \times I'_n,
 \end{equation}
where the $L$ is the loss function defined as in \autoref{eq:loss}. We generate multiple \subsetCF\ for each feature to analyze their effects on the model's predictions to the subgroup.
To speed up the generation of CF examples, we adapt the minibatch method\del{ to our optimization process}.

As such, we are able to support the \del{previously described }exploratory procedure for refining hypotheses.
If, in all \subsetCF\ groups, every CF example falls outside the feature ranges, the robustness of this claim is supported by even potentially unseen examples---even when we intentionally seek negative examples that could invalidate the hypothesis, it is not possible to do so. 
Otherwise, users may refine or reject the hypothesis as suggested by the \subsetCF\ groups (\autoref{fig:subgroup}C).


%% file: content/system.tex
\section{\system}

In this section, we first introduce the architecture and workflow of \system{}. Then we describe the design choices in the two main system interface components: \textit{table view} and \textit{instance view}.  \looseness=-1

\subsection{Overview}
As a decision exploration tool for machine learning models, \system{} is designed as a sever-client application.  
To make the system extensible with different models and counterfactual explanation algorithms, we design \system{} with three major components: the \emph{Data Storage} module, the \emph{CF Engine} module, and the \emph{Visual Analysis} module. 
\revise{}{The former two are integrated into a web-server implemented using Python with Flask, and the last one is implemented with React and D3\del{ as a few reusable front-end components, consolidated as an interactive visualization system}.}

The Data Storage module provides configuration options so advanced users can easily supply their own classification models and datasets. 
The CF Engine implements a set of algorithms for generating CF examples with fully customizable constraints. It also implements the procedure for producing subgroup-level CF examples (as described in \autoref{sec:subgroup-method}). 
The Visual Analysis module consists of an \textit{instance view} and a \textit{table view}. The \textit{instance view} allows a user to customize and inspect the CF examples of a single instance of interest (\ref{r:instance}, \ref{r:customize}). The \textit{table view} presents a summary of the instances of the dataset and their CF examples (\ref{r:subgroup}). It allows subgroup-level analysis through easy subgroup creation (\ref{r:selection}) and counterfactual comparisons (\ref{r:comparison}).  \looseness=-1

\begin{figure}
    \centering
    \includegraphics[width=\linewidth]{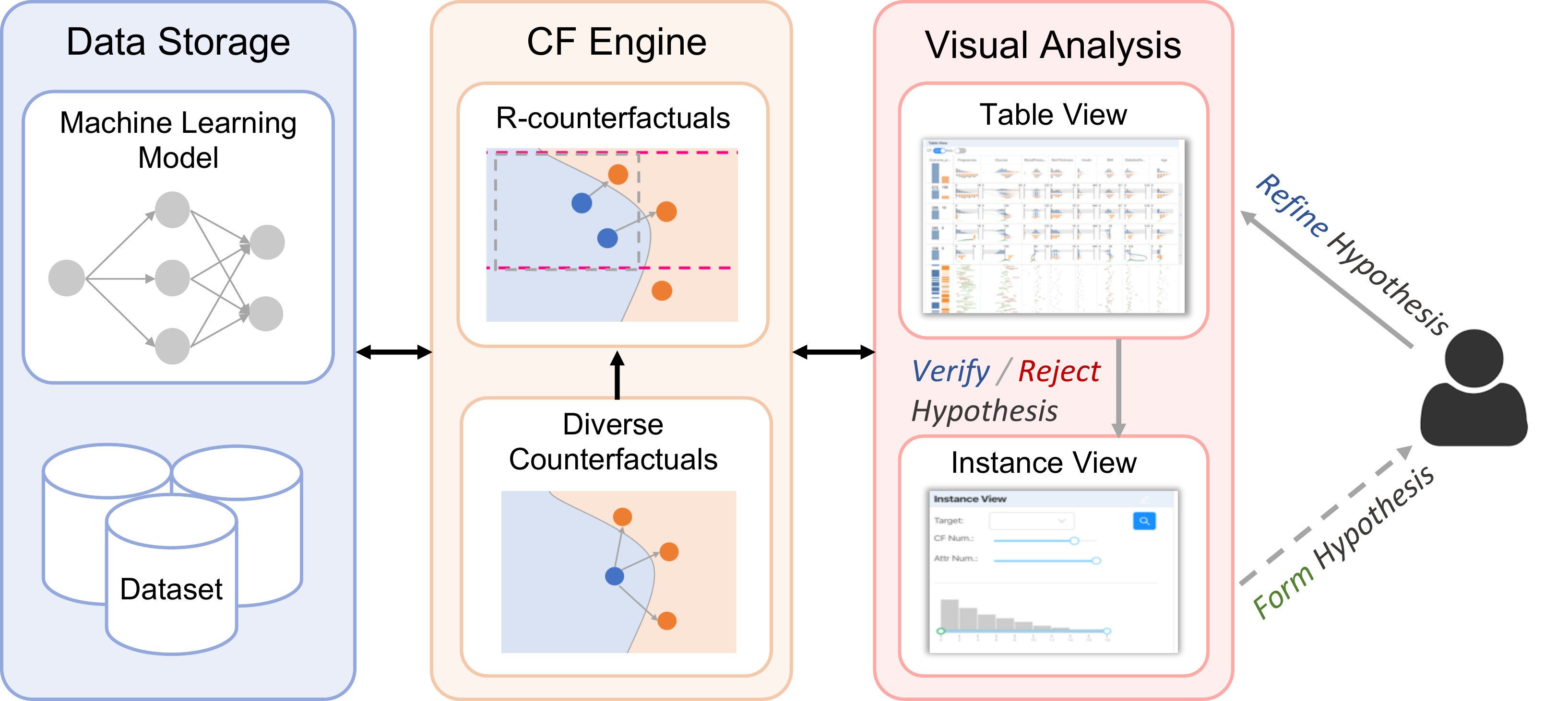}
    \vspace{-0.25in}
    \caption{\system{} consists of a \textit{Data Storage} module, a \textit{CF Engine} module, and a \textit{Visual Analysis} module.
    In the \textit{Visual Analysis} module,
    the \textit{table view} and \textit{instance view} together support an exploratory analysis workflow. \looseness=-1
    }
    \label{fig:arch}
    \vspace{-0.15in}
\end{figure}

The \textit{instance view} and \textit{table view} complement each other as a whole in supporting the exploratory analysis of model decisions. As shown in \autoref{fig:arch}, by exploring the diverse CF examples of specific instances in the \textit{instance view}, \revise{we}{users} can spot potentially interesting counterfactual phenomena and suggest related hypotheses. With hypotheses in mind, either formulated through exploration or prior experiences, \revise{we}{users} can utilize the r-counterfactuals integrated into the \textit{table view} to assessing plausibility. After refining a hypothesis, \revise{we}{users} can then verify or reject it by attempting to validate the corresponding instances in the \textit{instance view}.  \looseness=-1

\subsection{Visualization of Subgroup with R-counterfactuals}
\label{sec:subgroup-visualization}

Verifying and refining the hypothesis on the model's subgroup-level behaviors is the most critical and challenging part of the exploratory analysis procedure. We focus on one feature at a time and refine the hypothesis achieved with \subsetCF{} (described in \autoref{sec:subgroup-method}). 
For each \subsetCF{} group, we use a set of hybrid visualization to summarize and compare the \subsetCF{} and original subgroup's instance value on each focusing feature.   \looseness=-1

\begin{figure}

    \centering
    \includegraphics[width=\linewidth]{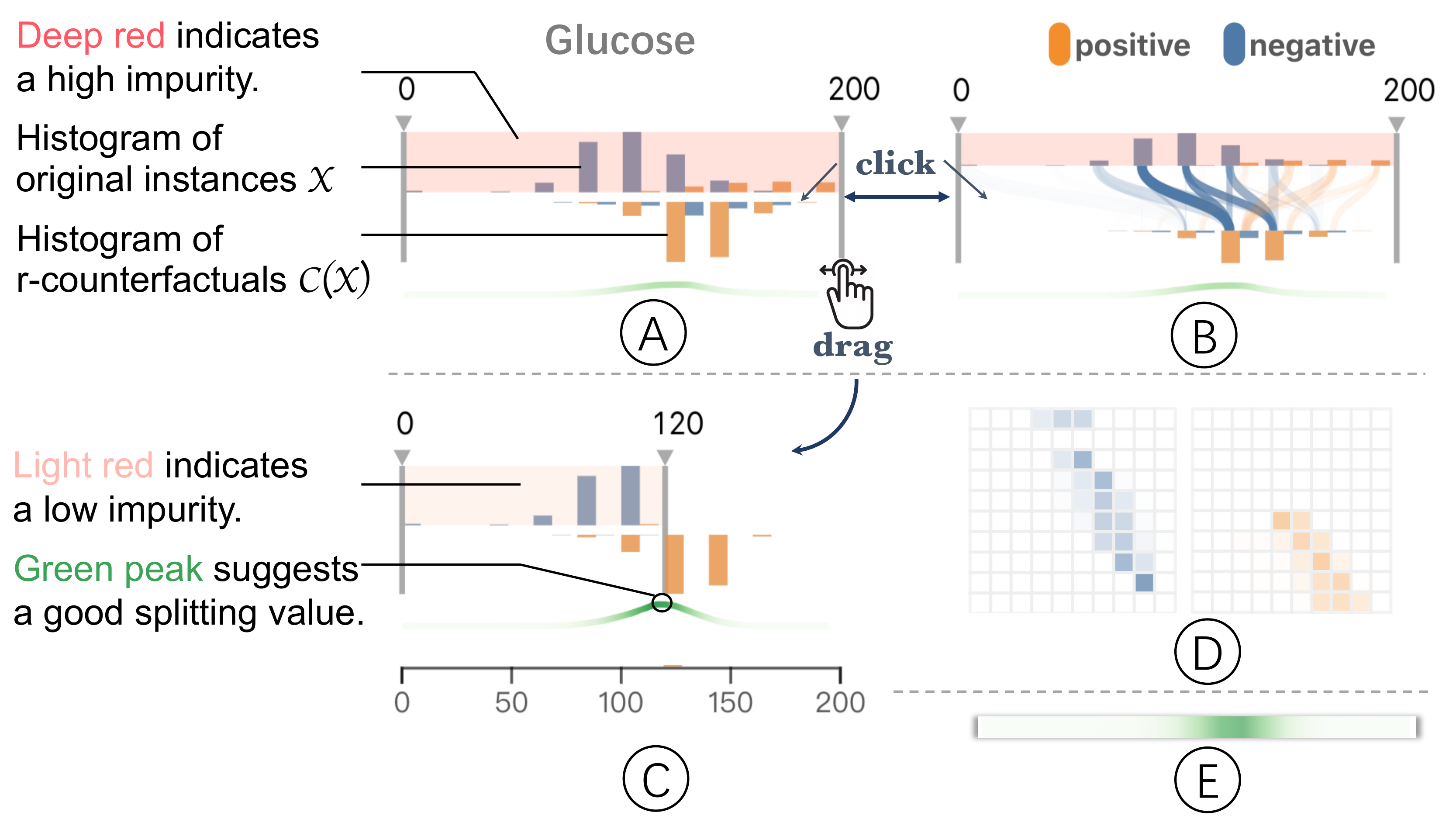}
        \vspace{-0.3in}
    \caption{Design choices for visualizing subgroup with r-counterfactuals. A-C. Our final choices.
    D. A matrix-based alternative design to visualize instance-cf connections.  
    E. A density-based alternative design to visualize the distribution of the information gain.}
    \label{fig:subgroup_design}
        \vspace{-0.2in}
\end{figure}

\textbf{Visualize Distribution (\ref{r:subgroup}).} To summarize an \subsetCF{} group, we use two side-by-side histograms to visualize the distributions of the original group and counterfactuals group (\autoref{fig:subgroup_design}A). The color of the bars indicates the prediction class. 
Sometimes, the system cannot find valid CF examples, which indicates that the prediction for these instances can hardly be altered by changing their feature values with the constraints hold. 
In this case, we use grey bars to indicate their number and stack them on the colored bar.  
The two histograms are aligned horizontally, with the upper one referring to the original data and the bottom one referring to the counterfactuals group. For continuous features, a shadow box with two handles is drawn to show the range of the subgroup's feature value. 
To provide a visual hint for the quality of the subgroup, we use the color intensity of the box to signify the Gini impurity of data in this range. 
\revise{}{When users select a subgroup containing instances all predicted as positive, a darker color indicates that (negative) CF examples can be easily found within this subgroup.}
In this case, the hypothesis might not be favorable and needs to be refined or rejected.
For categorical features, we use bar charts instead. Triangle marks are used to indicate the selected categories. \looseness=-1

\textbf{Visualize Connection.} CF examples are paired with original instances. The pairing information between the two groups can help users understand how the original instances need to change to flip their predictions.
In another sense, it also helps to understand how the local decision boundary can be approximated \cite{mothilal2020explaining}. 
Intuition hints that the feature with a larger change is likely to be a more important one. The magnitude of the changes also indicates how difficult the subgroup predictions are to flip by modifying that feature.
We use a Sankey diagram to display the flow from original group bins (as input bins) to counterfactual group bins (as output bins) (\autoref{fig:subgroup_design}B). 
For each link, the opacity encodes the flow amount while the width encodes the relative flow amount to the input bin size. 
An alternative design is the use of a matrix to visualize the flow between the bins (\autoref{fig:subgroup_design}D). Each cell in the matrix represents the number of instance-CF pairs that fail in the corresponding bins.
However, in practice, the links are quite sparse. Thus, compared with the matrix, the Sankey diagram saves much space and also emphasizes the major changes of CF from the original value, so we choose the Sankey diagram as our final design.   \looseness=-1

\textbf{Refine Subgroup (\ref{r:selection}).} As we have discussed in \autoref{sec:subgroup-method}, a major task during the exploration is to assess and refine the hypothesis. What is a plausible hypothesis represented by a subgroup? A general guideline is: a subgroup is likely to be a good one if it separates instances with different prediction classes. The intuition is that if a boundary can be found to separate the instances, the refined hypothesis is likely to be valid.
We choose the information gain based on Gini impurity to indicate a good splitting point, which is widely used in decision tree algorithms \cite{breiman1984classification}.
The information gain is computed as $1 - (N_{left}/N)\cdot  I_G(D_{left}) - (N_{right}/N)\cdot I_G(D_{right})$, where $D_{left}, D_{right}$ are the data including both original instances and CF examples, split into the left set or the right set. 
At first, we try to visualize the information gain in a heatmap lie between the two histograms (\autoref{fig:subgroup_design}E). 
Though such design has the advantage of not requiring extra space, we find that it is hard to distinguish the splitting point with the maximum information gain from the heat map.
To make the visual hint salient, we visualize the impurity scores as a small Sparkline under the histograms. 
We use color and height to double encode this information.
For continuous features, we enable users to refine the hypothesis by dragging the handles to a new range. And for categorical features, users are allowed to click the bars to update the selected categories.  \looseness=-1 

\subsection{Table View}
The \textit{table view} (Fig.\ref{fig:teaser}A) is a major component and entry point of \system. Organized as a table, this view summarizes the subgroups with their \subsetCF\, as well as the details of instances in a focused subgroup. 
Vertically, the \textit{table view} consists of three parts. From top to bottom, they are the table header, the subgroup list, and the instance lens. 
The table header is a navigator to help users explore the data in the rest of the table. The subgroup list shows a summary of multiple groups, while the instance lens shows details for a specific subgroup. The three components are aligned horizontally with multiple columns, each corresponding to a feature in the dataset. The first column, as an exception, shows the label and prediction information. 
Next, we explain the three parts and interactions that enable them to work together.   \looseness=-1

\textbf{Table Header.} 
The table header (Fig.\ref{fig:teaser}A1) presents the overall data distribution of the features in a set of linked histograms/bar charts. The first column shows the predictions of the instances in a stacked bar chart, where each bar represents a prediction class. Users can click a bar to focus on a specific class. \revise{}{We indicate false predictions by a hatched bar area and true predictions by a solid bar area.}
In each feature column, we also use two histograms/bar charts (introduced in \autoref{sec:subgroup-visualization}) to visualize the distribution of the instances and CF examples. Users can efficiently explore the data and their CF examples via filtering and sorting. 
For each feature, filtering is supported by brushing (for continuous features) or clicking (for categorical features) on both histograms/bar charts. After filtering, that data is highlighted in the histograms while the rest is represented as translucent bars. To sort, users can click on the up/down buttons that pop-up when hovering on each feature.  \looseness=-1

\textbf{Subgroup List.} 
The subgroup list (Fig.\ref{fig:teaser}A2) allows users to create, refine, and compare different subgroups. Here, each row corresponds to a subgroup. 
The first column presents the predictions of the instances in the same design used in the table header. 
In other columns, each cell $(i, j)$ presents a summary of the subgroup $i$ with \subsetCF\ for the feature $j$ (introduced in \autoref{sec:subgroup-visualization}). 
The subgroup list is initialed with one default group, which is the whole dataset with unconstrained CF examples. 
Starting with this group, users can refine the group by changing ranges for each feature and clicking the update button. 
\revise{Users can also copy a subgroup for stashing or delete an unwanted subgroup.}{The users can also copy or delete an unwanted subgroup to maintain the subgroup list, which helps track explorations history.}
By aligning different subgroups together, users can gain insights from a side-by-side comparison (\ref{r:comparison}). 


\textbf{Instance Lens.} 
When users click a cell in the subgroup list, the instance lens (Fig.\ref{fig:teaser}A3) presents details pertaining to each instance and CF examples. To ensure the scalability to a large group of instances, we design an instance lens based on Table Lens \cite{rao1994table}, a technique for visualizing large tables. In the instance lens, we design two types of cells: row-focal and non-focal cells with different visual representations. \revise{}{A CF example shows minimal changes from the original instance. For numerical features, we use a line segment to show how the change is made. The x-position of the two endpoints encodes the feature value of the original instance and the CF example. The endpoint corresponding to the original instance is marked black.} We use green and red to indicate a positive or negative change. For categorical features, \revise{the position of two lines is used to encode the category of the original instance and CF example, where the line with a deeper color represents the original value.}{we use two broad line segments to indicate the category of the original instance (in a deeper color) and the CF example (in a lighter color).} Users can focus on a few instances by clicking them. Then the non-focal cells become row-focal cells where the text of the exact feature value is presented.   \looseness=-1

\subsection{Instance View}

The \textit{instance view} helps users inspect the diverse counterfactuals of a single instance (\ref{r:instance}). The inspection results can be used to support or reject their hypothesis formed from the \textit{table view}. 
For users, mostly decision subjects \textit{instance view} can be used independently to find actionable suggestions to achieve the desired outcome.  \looseness=-1

The \textit{instance view} consists of two parts. The setting panel (\autoref{fig:teaser}B1) shows the prediction result of the input instance as well as the target class for the CF examples. In the panel, users can also set the number of CF examples they expect to find and the maximum number of features that are allowed to change to ensure the sparsity. 
The resulting panel (\autoref{fig:teaser}B2) displays each feature in a row. It allows users to input a data instance by dragging sliders to set values of each feature.
The distribution of each feature value is presented in a histogram, which suggests the users compare the instance's feature values with the overall distribution of the whole dataset. 
For each feature, uses can add constraints (\ref{r:customize}) by setting the value range of CF examples or locking the feature to prevent it from changing. \revise{}{The interactions help decision subjects to customize actionable counterfactual explanations for their scenarios.}
\del{Users can also change the order of the features by clicking a ``down'' button. Then the feature will be put into the bottom of the list so that users could keep a few important features on the screen.}   \looseness=-1

After users input the instance, they click the ``search'' button in the panel.
The system then returns all CF examples found in both valid and invalid sets. 
The CF examples, as well as the original instances, are presented in a set of polylines along parallel axes (\autoref{fig:teaser}B2), where we use color to indicate their validity and prediction class. 
For valid CF examples and the original instance, we apply the same color use in the \textit{table view} to show their prediction class, while invalid CF examples are presented in grey. 
Users can depict details of a valid CF example by hovering on the polyline. It is then highlighted, and the text of each feature value will be presented on the axis.   \looseness=-1

%% file: content/evaluation.tex
\section{Evaluation}
\label{sec:evaluation}
Next, we demonstrate the efficacy of \system{} through three usage scenarios targeting three types of users: decision-makers, model developers, and decision subjects. We also gather feedback from expert users through formal interviews and interactive trails of the system.  \looseness=-1

\subsection{Usage Scenario: Diabetes Classification} 
\label{sec:diabetes}

\begin{figure*}
    \centering
    \includegraphics[width=0.9\textwidth]{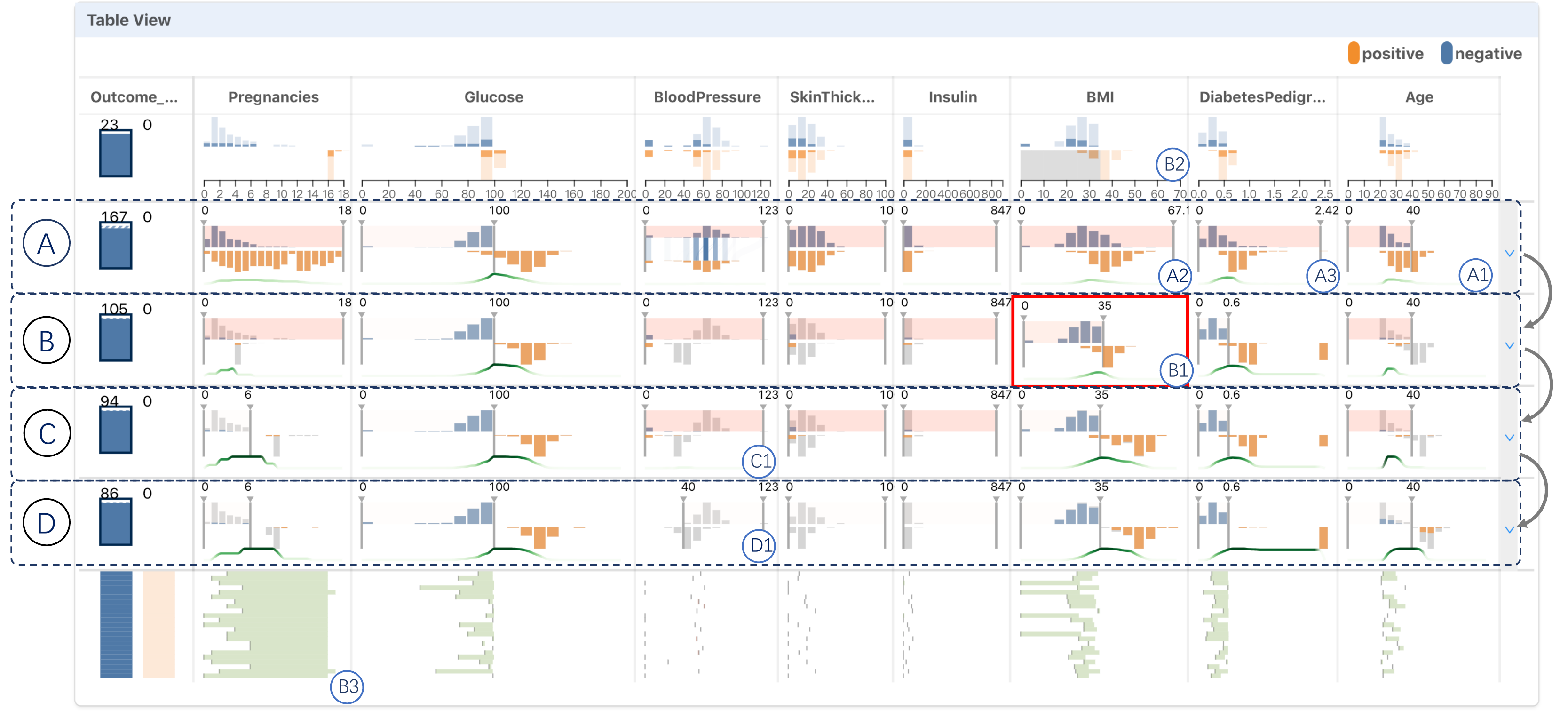}
     \vspace{-0.15in}
    \caption{Understanding diabetes prediction on a young subgroup with a hypothesis refinement process. A. The initial hypothesis---``patients with ${Glucose < 100}$ and ${age < 40}$ are healthy''---is not valid as suggested by several CF examples found inside the range (A1). B. The hypothesis is refined by constraining $BMI < 35$, and $DPF < 0.6$, as suggested by the green peaks under the histograms (A2, A3). The hypothesis is still not plausible given the CF examples within the range (B1). After studying cell B1, all CF examples have an uncommonly high pregnancy value (B3). C. The user refines the hypothesis by limiting $pregnancies <= 6$. All valid CF examples left in the subgroup have abnormal blood pressure (C1). D. After filtering out instances with an abnormal low ($< 40$) blood pressure (D1), the final hypothesis is now fully supported by all CF examples.
    }
    \label{fig:case-1}
     \vspace{-0.20in}
\end{figure*}


In the first scenario, Emma, a medical student who is interested in the early diagnosis of diabetes, wants to figure out how a diabetes prediction model makes predictions.
She has done some medical data analytics before, but she does not have much machine learning knowledge. 
She finds the Pima Indian Diabetes Dataset (PIDD) \cite{smith1988using} and downloads a model trained on PIDD. 
The dataset consists of medical measurements of 768 patients, who are all females of Pima Indian heritage and at least 21 years old. The task is to predict whether a patient will have diabetes within 5 years. The features of each instance include the number of pregnancies, glucose level, blood pressure, skin thickness, insulin, body mass index (BMI), diabetes pedigree function (DPF), and age, which are all continuous features. 
The dataset includes 572 negative (healthy) instances and 194 positive (diabetic) instances. 
The model is a neural network with two hidden layers with 76\% and $79\%$ test and training accuracy, respectively.   \looseness=-1

\textbf{Formulate Hypothesis (\ref{r:selection}). }
Emma is curious about how the model makes predictions for patients younger than 40. 
Based on her prior knowledge, she formulates a hypothesis that patients with $\textup{age} < 40$ and $\textup{glucose} < 100\ \textup{mmol/L}$ are likely to be healthy. 
She loads the data and the model to DECE and creates a subgroup by limiting the ranges on age and glucose accordingly.  \looseness=-1

\textbf{Refine Hypothesis. }
The subgroup consists of 167 instances, all predicted as negative. However, she notices that for most of the features, the range boxes are colored with a dark red (\autoref{fig:case-1}A), indicating that CF examples can be found within the subgroup. 
This suggests that patients with a diabetic prediction potentially exist in the subgroup and implies that the hypothesis may not be valid. 
After a deeper inspection, she finds a number of CF examples (\autoref{fig:case-1}A1) with age $<$ 40.
Then she checks each feature and tries to refine the hypothesis by restricting the subgroup to a smaller one.
She finds that the BMI distribution of CF examples is shifted considerably to the right in comparison to the original distribution  (\autoref{fig:case-1}A2). This means that BMI needs to be increased dramatically to flip the model's prediction to positive.
Thus, Emma suspects that the original hypothesis may not hold for patients with high BMI (obesity).
She proceeds to refine the subgroup by adding a constraint on BMI. The green peek in the bottom sparkline (\autoref{fig:case-1}A2) suggests that a split at $BMI = 35$ could make a good subgroup. In the meantime, she discovers a similar pattern for DPF (\autoref{fig:case-1}A3). She adds two constraints of $BMI < 35$ and $DPF < 0.6$, and then she creates an updated subgroup.   \looseness=-1

After the refinement, Emma finds that most of the cells are covered by a transparent range box or filled with mostly light grey bars (\autoref{fig:case-1}B). The light grey area represents the instances for which no valid CF examples can be found. One exception is the BMI cell (\autoref{fig:case-1}B1), where a few CF examples can still be generated. 
To check the details, Emma focuses on this subgroup by clicking the ``zoom-in'' button. After the table header is updated, she filters the CF examples with $BMI < 35$ by brushing on the header cell (\autoref{fig:case-1}B2). 
She finds that all the valid CF examples have an extreme pregnancy value (\autoref{fig:case-1}B3), which means patients \revise{with a high pregnancy}{who have had several pregnancies} are exceptions to the hypothesis.
However, such pregnancy values are very rare, so Emma updates the subgroup with a constraint of $pregnancy < 6$, which covers most of this subgroup.   \looseness=-1

After the second update, the refined hypothesis is almost valid, yet some CF examples can still be found in the subgroup, which challenges the hypothesis (\autoref{fig:case-1}C1). By checking the detailed instances of ``unsplit'' feature groups, she finds that all these valid CF examples have an unlikely blood pressure of 0. These blood pressure values are likely caused by missing data entries. She raises the minimum blood pressure value to a normal value of 40 and then makes the third update. 
The final subgroup consists of 86 instances, all predicted negative for diabetes, and all feature cells are covered with a totally transparent box, indicating a fully plausible hypothesis. \looseness=-1

\textbf{Draw Conclusions.}
After three rounds of refinement, Emma concludes that patients with $age < 40$, $glucose < 100,\ BMI < 35,\ \textup{and}\ DPF < 0.8$ are extremely unlikely to have diabetes within the next five years, as suggested by the model. 
This statement is supported by 86 out of 768 instances in the dataset (11\%) and their corresponding counterfactual instances.
Generally, Emma is satisfied with the conclusion but concerned with something she sees in the blood pressure column. It seems that although all the bars turn grey, a clear shift between the two histograms exists (\autoref{fig:case-1}D1), indicating that the model is trying to find CF examples with low blood pressures.  \looseness=-1

She is confused about why low blood pressure can also be a symptom of diabetes. 
Initially, she thinks it may be a local flaw in the model caused by mislabeled instances of 0 blood pressure. 
So she finds another model trained on a clean dataset and runs the same procedure. However, the pattern still exists.
After some research,  she finds out that a diabetes-related disease called Diabetic Neuropathy\footnote{https://en.wikipedia.org/wiki/Diabetic\_neuropathy} may cause low blood pressure by damaging a type of nerve called autonomic neuropathy. She concludes the exploratory analysis and gains new knowledge from the process.  \looseness=-1


\begin{figure*}
    \vspace{-0.25in}
    \centering
    \includegraphics[width=\textwidth]{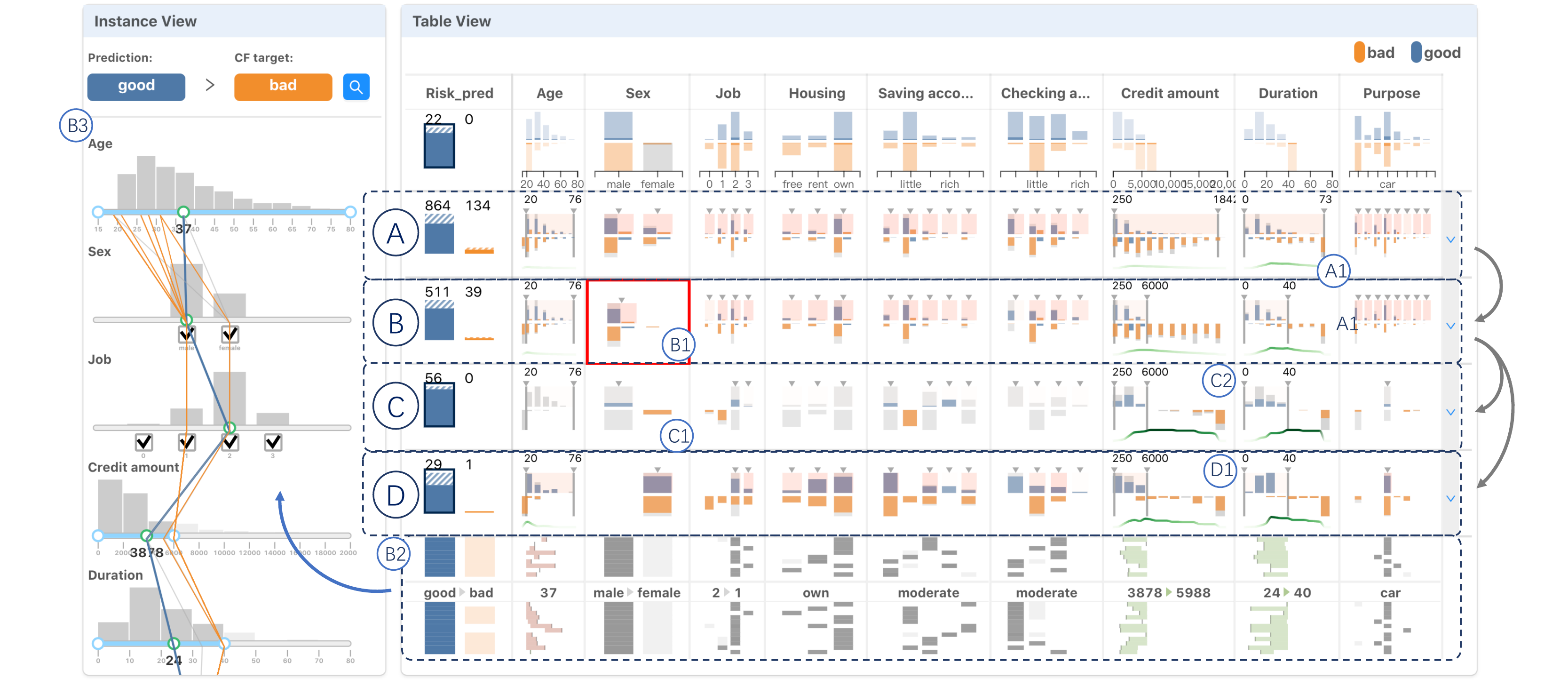}
     \vspace{-0.25in}
    \caption{Subgroup comparison on a neural network trained on German Credit Dataset. 
    A. The whole dataset with CF examples where the distribution of the credit duration (A1) suggests that a long duration of debt will lead to a ``bad'' risk for all loan applicants. 
    B. The male subgroup that covers a majority of male instances, where the gender column (B1) suggests that a few CF examples are generated by changing the gender from male to female (B2). 
    B3. Diverse CF examples against a sample male instance from B2, where all valid CF examples (orange lines) suggest either to change the gender from male to female or to degrade the job rank.
    C. The narrowed male subgroup where a larger portion of the instances have CF examples that change their gender to female (C1). The CF examples are found far from the subgroup (C2).
    D. The contrast female subgroup against the male subgroup C, where CF examples can be found within the subgroup (D1). \looseness=-1
    }
    \label{fig:case-2}
     \vspace{-0.2in}
\end{figure*}

\subsection{Usage Scenario: Credit Risk}
In this usage scenario, we show how \system{} can help model developers understand their models. Specifically, we demonstrate how comparative analysis of multiple subgroups can support their exploration.\looseness=-1

A model developer, Billy, builds a credit risk prediction model from the German Credit Dataset \cite{hofmann1994statlog}. The dataset contains credit and personal information of 1000 people and their credit risk label of either good or bad. The model he trained is a neural network with two hidden layers, which achieves an accuracy of 82.2\% on the training set and 74.5\% on the test set. Billy wants to know how the model makes predictions for different subgroups of people (\ref{r:comparison}). Particularly, he wants to figure out if gender affects the model's prediction and the behavior of the model's predictions on different gender groups. 

Billy begins the exploration by using the whole dataset and CF examples (\autoref{fig:case-2}A). 
He notices that most of the CF examples with a flipped prediction as bad-risk have an extremely large amount of debt or debt that has lasted a long time (\autoref{fig:case-2}A1). This finding indicates that almost certainly, a client with an extremely large credit amount or duration would be predicted as a bad credit risk.   \looseness=-1

\textbf{Select Subgroup (\ref{r:selection}).} 
He wants to learn if there are other factors that might affect credit risk prediction.
So he creates a subgroup with $credit\ amount < 6000\ \textup{DM}\ \textup{and}\ duration < 40\ \textup{months}$, containing a majority of the dataset. 
Then he selects the male group to inspect the model's predictions and counterfactual explanations.
The male group (\autoref{fig:case-2}B) contains a total of 550 instances, and 511 instances are predicted to be good candidates. 
Billy checks the gender column and finds that a few CF examples are generated by changing the gender from male to female (\autoref{fig:case-2}B1). 
After Billy zooms in to the gender cell and filters the CF examples with $gender=female$ (\autoref{fig:case-2}B2), he finds that 22 instances have CF examples with gender changed. This indicates that gender may be considered in the prediction.   \looseness=-1

\textbf{Inspect Instance. }  Billy wants to figure out whether this pattern is intentional or accidental. He randomly puts one instance into the \textit{instance view}\del{ to check how the model makes the tradeoffs between different feature values for altering the prediction from ``good'' to ``bad''}. \revise{}{Billy sets the feature ranges, $credit\ amount < 6000\ \textup{and}\ duration < 40$ to be the same as the subgroup's feature ranges. He clicks the search button to find a set of CF examples to probe the model's local decision (\autoref{fig:case-2}B3). He finds that all valid CF examples alter the prediction from good credit candidates to bad credit candidates by either suggesting a worse job (down by one rank) or changing the gender from male to female. 
He locks the job attribute and tries again. He finds that all CF examples suggest changing the gender from male to female.}
This indicates that gender affects the model's prediction in this instance and similar ones.  \looseness=-1

\textbf{Refine Subgroup.} \revise{}{Then Billy wants to find if there are any subgroups where gender has an enormous impact. He finds most of the male instances with unactionable CF examples that need to change their genders are from a subgroup of wealthy people, who have a good job (2-3), an above-moderate saving account balance, and apply for credit in order to buy a car (\autoref{fig:case-2}B2).} Billy creates a wealthy subgroup and makes the update (\autoref{fig:case-2}C). Then he finds that compared with the former group, a larger portion of the instances have CF examples that change their gender to female (\autoref{fig:case-2}C1). The major shift of gender in the CF examples implies that gender plays a more important role in the predictions of the wealthy subgroup.  \looseness=-1

\textbf{Compare Subgroups (\ref{r:comparison}). }To confirm this observation, Billy compares the male subgroup with another female subgroup, which has all of the same feature ranges except gender (\ref{r:comparison}). 
Billy finds that all 56 instances in the male subgroup are predicted as good credit candidates, and in the female subgroup, 29 out of 30 instances are predicted to be good credit candidates (\autoref{fig:case-2}D).
Then Billy compares the distribution of the CF examples against the two subgroups. 
In the male subgroup (\autoref{fig:case-2}C), indicated by the transparent range boxes, all the valid CF examples are generated out of range. These CF examples support the conclusion that any potential male applicant within this subgroup is very unlikely to be predicted as a bad credit risk. In the credit amount column (\autoref{fig:case-2}C1), all the valid CF examples are found with $credit\ amount > 10000\ \textup{DM}$. This means that if a man, who has a good job (2-3) and an above-moderate saving accounts balances, applies for a line of credit of 10000 DM with a term of fewer than 40 months, it is very likely that his request will be approved.
However, in the female subgroup (\autoref{fig:case-2}D), a large amount of CF examples are found within this group. In the credit amount column (\autoref{fig:case-2}D1), CF examples can be found with $credit\ amount < 6000 \ \textup{DM}$. This indicates that a loan request of $credit\ amount = 6000\ \textup{DM}$ from a woman in the same condition may be rejected. \looseness=-1

\textbf{Draw Conclusions. }After the exploratory analysis, Billy concludes that gender plays an important role in the model's prediction for a subgroup of wealthy people defined above. 
In addition, the finding is supported by both instance-level and subgroup-level evidence. 
This is a strong sign that the model might behave unfairly towards different genders. Billy then decides to inspect the training data and conduct further research to eliminate the model's probable gender bias.  \looseness=-1

\subsection{Usage Scenario: Graduate Admissions}

In this usage scenario, we show how \system{} provides model subjects with actionable suggestions. Vincent, an undergraduate student with a major in computer science, is preparing an application for a master's program. He wants to know how the admission decision is made and if there are any actionable suggestions that he can follow. 
He has downloaded a classifier\revise{}{, a neural network with two hidden layers} trained on a Graduate Admission Dataset \cite{acharya2019comparison} to predict whether a student will be admitted to a graduate program. He inputs his profile information and the model predicts that he is unlikely to be accepted. He wants to know how he can improve his profile to increase his acceptance chance.   \looseness=-1

\begin{figure}
    \vspace{-0.05in}
    \centering
    \includegraphics[width=\linewidth]{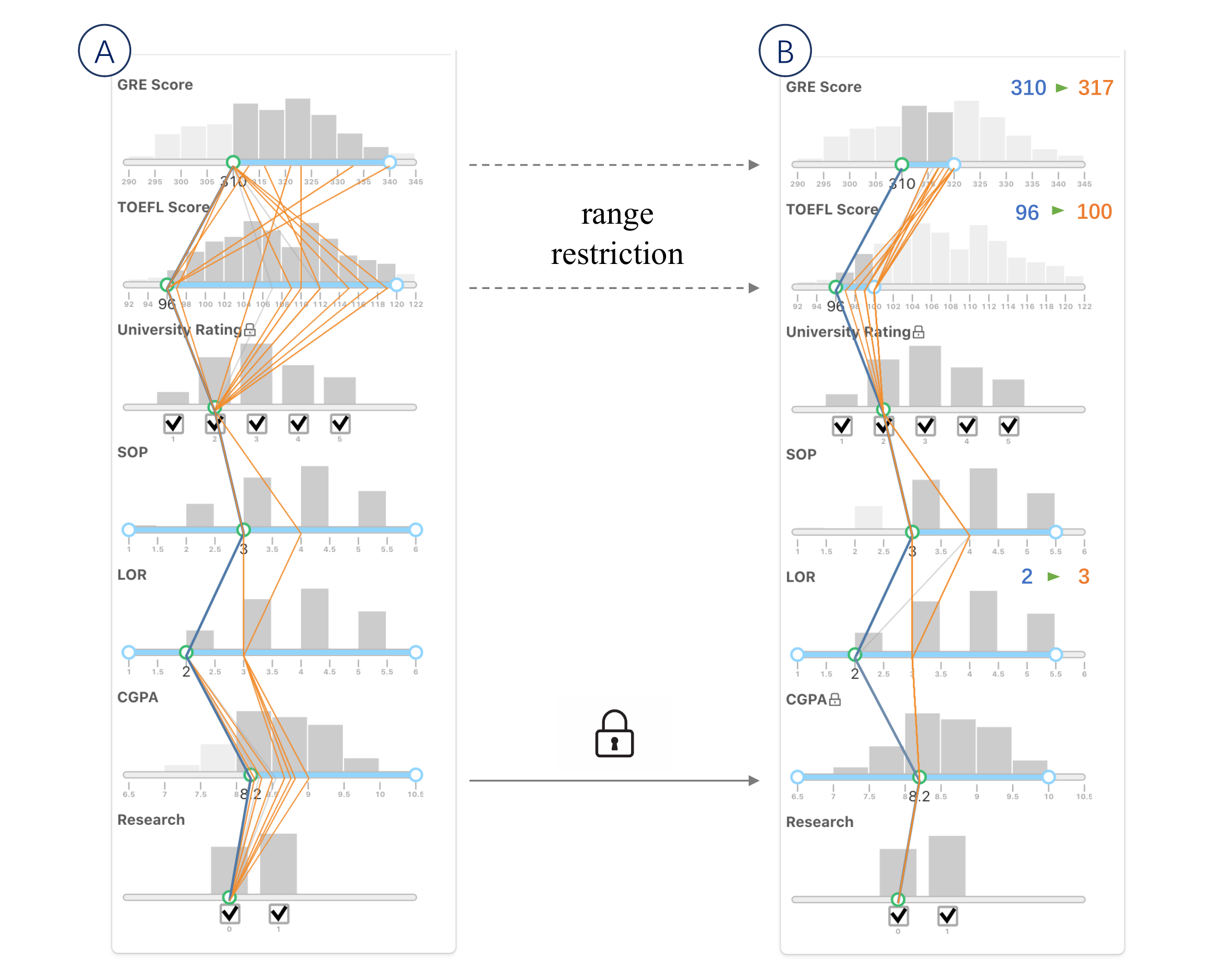}
    \vspace{-0.30in}
    \caption{A student gets actionable suggestions for graduate admission applications. A. Diverse CF examples generated with university rating locked. B. Customized CF examples with user constraints, including $\text{GRE score} < 320$, $\text{TOFEL score} < 100$, and a locked CGPA.}
    \label{fig:case-3}
     \vspace{-0.20in}

\end{figure}

He chooses DECE and focuses on the \textit{instance view}. He inputs his profile information using the sliders. He finds that compared with other instances in the whole dataset, his cumulative GPA (CGPA) of 8.2 is average while both his GRE score, 310, and TOFEL score, 96, are below average. According to the provided rating scheme, he sets his undergraduate university rating, statement of purpose, and recommendation letter as 2, 3, and 2, respectively, which are all below average. 
Then he attempts to find valid CF examples. 
First, he goes through the attributes and locks the university rating since it is not changeable (\ref{r:customize}). Then he sets the number of CF examples to 15 and clicks the search button to get the results (\autoref{fig:case-3}A). He is surprised that so many diverse CF examples can be found (\ref{r:instance}). However, he finds that over half of these CF examples have either a high GRE score (above 320) or TOEFL score (above 108), which would be challenging for him to achieve. 
Since he has already finished most of the courses in the undergraduate program, it would be difficult to boost his CGPA.
So, considering his current situation, he locks the CGPA as well and tries to find CF examples with $GRE<320$ and $TOEFL<100$ (\ref{r:customize}). 
In the second round of attempts, the system can still find a few valid CF examples (\ref{fig:case-3}B). He notices that all valid CF examples contain a GRE score above 315, which might indicate a lower bound of the GRE score initially suggested by the model. Also, all the valid CF examples suggest that a stronger recommendation letter would help. \looseness=-1

Finally, Vincent is satisfied with the knowledge learned from the investigation and decides to pick a CF example to guide his application preparation, which suggests he increase his GRE score to 317, TOEFL score to 100, and obtain a stronger recommendation letter.   \looseness=-1

\subsection{Expert Interview}
\label{expert-interview}

\revise{}{To validate how effective \system{} is in supporting decision-makers and decision subjects, we conduct individual interviews with three medical students (E1, E2, E3). All medical students have had a one-year clinical internship in hospitals. They have basic knowledge of statistics and can read simple visualizations. These (almost) medical practitioners can be regarded as decision-makers in the medical industry. Also, they have experience interacting with patients and can offer valuable feedback regarding the need from end-users.

The interviews were held in a semi-structured format. We first introduced the counterfactual-based methods by cases and figures without algorithm details. Then we introduced the interface of \system{} using the live case example from \autoref{sec:diabetes}. The introductory session took 30 minutes. Afterward, we asked them to explore \system{} using the Pima Indian Diabetes Dataset for 20 minutes and collected feedback about their user experience, scenarios for using the system, and suggestions for improvements.


\textbf{System Design.}
Overall, E1 and E2 suggested that the system is easy to use, while E3 had some confusion with the instance lens, where she mistakenly thought that the line in a non-focal cell encodes some range values. After shown with more concrete examples about how the non-focal cells and row-focal cells switch to each other, she got to understand. 
E1 mentioned that the color intensity of the shadow box in each subgroup shows cells in a very prominent way, suggesting a group of healthy patients have potential risks to have the disease. 

\textbf{Usage Scenarios.}
E1 suggested that the subgroup list in the \textit{table view} helped her understand the potential risk level of her patients becoming diabetic, ``\textit{so that I can decide if any medical intervention should be taken.}'' However, both E2 and E3 suggested that the \textit{table view} would be more useful in clinical research scenarios, such as understanding the clinical manifestations of multi-factorial diseases.\del{ E3 gave an example of complex diagnosis of Systemic Lupus Erythematosus (SLE).} ``\textit{Doctors can then use the valid conclusions for diagnosis},'' E2 said. When asked whether they would recommend patients use the \textit{instance view} to find suggestions by themselves, E2 and E3 showed concerns, saying that ``\textit{the knowledge variance of patients can be very broad}.'' Despite this concern, they mentioned that they would love to use \system{} themselves to help patients find disease prevention suggestions.

\textbf{Suggestions for Improvements.}
All three experts agreed that the \textit{instance view} has a high potential for doctors and patients to apply it in clinical scenarios together, but some domain-specific problems must be considered. E1 suggested that we should allow users to set the value of some attributes as not applicable because ``\textit{for the same disease, the test items can be different for patients}.'' E2 commented that the histograms for each attribute did not help her much. She suggested showing the reference ranges for each attribute as another option. 
}
\looseness=-1

%% file: content/discussion.tex
\section{Discussion and Conclusion}

\revise{}{In this work, we introduced \system{}, a counterfactual explanation approach, to help model developers and model users explore and understand ML models' decisions. \system{} supports iteratively probing the decision boundaries of classification models by analyzing subgroup counterfactuals and generating actionable examples by interactively imposing constraints on possible feature variations. We demonstrated the effectiveness of \system{} via three example usage scenarios from different application domains, including finance, education, and healthcare, and included an expert interview.}

\revise{}{
\textbf{Scalability to Large Datasets.}
The current design of the \textit{table view} can visualize nine subgroup rows or about one thousand instance rows (one instance requires one vertical pixel) on a typical laptop screen with 1920 $\times$ 1080 resolution. 
Users can collapse the features they are not interested in and inspect the details of a feature by increasing the column width. In most cases, about ten columns can be displayed in the table, as shown in the first and second usage scenarios.

In addition to the three datasets used in \autoref{sec:evaluation}, we also validated the efficiency of our algorithm on a large dataset, HELOC\footnote{https://community.fico.com/s/explainable-machine-learning-challenge} (10459 instances). We used a 2.5GHz Intel Core i5 CPU with four cores. Generating single CF examples for the entire dataset took around 16 seconds. We found that the optimization converges slower for datasets with many categorical attributes, such as the German Credit Dataset. To speed up the generation process, we envision future research to apply strategies from synthetic tabular data generation literature, such as smoothing categorical variables\del{ with softmax} \cite{xu2018synthesizing} and parallelism.}

\revise{}{
\textbf{Generalizability to Other ML Models.}
The \system{} prototype was developed for differentiable binary classifiers on tabular data. However, \system{} can be extended for multiclass classification by either enabling users to select a target counterfactual class or heuristically selecting a target class for each instance (\eg, the second most probable class predicted by the model). 
For non-differentiable models (\eg, decision trees) or models for unstructured data (\eg, image and text), generating good counterfactual explanations is an active research problem and we expect to support this in future.
}

\revise{}{
\textbf{R-counterfactuals and Exploratory Analysis.}
R-counterfactuals are flexible instruments to understand the behavior of a machine learning model in a subgroup. A general workflow for using it with \system{} is as follows: 1) users start by specifying a subgroup of interest; 2) from the subgroup visualization, users can view the class and impurity distribution along with each feature; 3) users can then choose an interesting or salient feature and further refine the subgroup; 4) continue 2) and 3) until they get a comparably ``pure'' subgroup, which implies that the findings on this subgroup are salient and valid.  
}

\revise{}{
\textbf{Improvements in the System Design.}
We expect to make further improvements to the system design in the future. In the \textit{table view}, we plan to improve the design of subgroup cells for categorical features. One direction is to provide suggestions for an optimal selection of categories to refine the hypothesis. In the \textit{instance lens}, we plan to implement more interactions to support the ``focus + context'' display (\eg, focusing through brushing). In the \textit{instance view}, we plan to provide customization options in different levels of flexibility for users to choose from, \eg, allowing advanced users to manipulate the feature weights in the distance metric directly.
}

\revise{}{
\textbf{Effectiveness of Counterfactual Explanations.}
In this paper, we presented three usage scenarios to demonstrate how \system{} can be used by different types of users. However, we only conducted expert interviews with medical students who can be regarded as decision-makers (for medical diagnosis). To better understand the effectiveness of \system{} and CFs, we plan to conduct user studies by recruiting model developers, data scientists, and layman users (for the \textit{instance view} only). At the instance-level, we expect to understand how different factors (\eg, proximity, diversity, and sparsity) affect the users' satisfaction with the CF examples. At the subgroup-level, an interesting research problem is how well r-counterfactuals can support exploratory analysis for datasets in different domains. 
}